\documentclass{article}

 \usepackage[preprint]{neurips_2026}

\usepackage[utf8]{inputenc} 
\usepackage[T1]{fontenc}    
\usepackage{hyperref}       
\usepackage{url}            
\usepackage{booktabs}       
\usepackage{amsfonts}       
\usepackage{nicefrac}       
\usepackage{microtype}      
\usepackage{xcolor}         
\usepackage{ulem}

\usepackage[most]{tcolorbox}
\usepackage{url}
\usepackage{graphicx}
\usepackage{subcaption}
\usepackage{soul}
\usepackage{graphicx}
\usepackage{xspace}
\usepackage{amsmath}
\usepackage{multirow}
\usepackage{tabularx}
\usepackage{colortbl}
\usepackage{wrapfig} 
\usepackage{pifont}
\usepackage{booktabs}
\usepackage{amsthm}   
\usepackage{amsmath}  
\usepackage{amssymb}
\usepackage{amsfonts} 
\usepackage{adjustbox}
\usepackage{enumitem}

\usepackage[table]{xcolor} 
\usepackage{colortbl}      

\usepackage{amsthm}

\newtheorem{proof}{Proof}

\definecolor{rowblue}{RGB}{228, 229, 252}
\definecolor{paperyellow}{RGB}{248,208,76}

\newcommand{\fakeparagraph}[1]{\noindent\textbf{#1.}}
\newcommand{\methodname}{{TaC-C}\xspace}

\newcommand\tabref[1]{Table~\ref{#1}}

\newcommand\appref[1]{Appendix~\ref{#1}}

\definecolor{lightbluepurple}{RGB}{235, 240, 250}

\usepackage{booktabs}
\usepackage{tabularx}
\usepackage{xcolor}
\usepackage{array}

\definecolor{extract}{HTML}{D6008F}
\definecolor{summary}{HTML}{0087C7}
\definecolor{assoc}{HTML}{007A78}
\definecolor{keyorange}{HTML}{E66A00}
\definecolor{softgray}{HTML}{555555}

\newcommand{\graytrace}[1]{\textcolor{softgray}{#1}}

\title{Thinking as Compression: Your Reasoning Model is Secretly a Context Compressor}

\author{%
\textbf{Guoxin Ma\textsuperscript{1,2,*} \quad
Yibing Liu\textsuperscript{1,*} \quad
Chengzhengxu Li\textsuperscript{2} \quad
Yu Liang\textsuperscript{1} \quad
Yan Wang\textsuperscript{1}}
\\
\textbf{Yueyang Zhang\textsuperscript{1} \quad
Kecheng Chen\textsuperscript{3} \quad
Zhaohan Zhang\textsuperscript{4} \quad
Zhiyuan Sun\textsuperscript{1} \quad
Daiting Shi\textsuperscript{1}}
\\[0.4em]
\textsuperscript{1}Baidu Inc., Beijing, China \quad
\textsuperscript{2}Xi'an Jiaotong University
\\
\textsuperscript{3}City University of Hong Kong \quad
\textsuperscript{4}Queen Mary University of London
\\[0.2em]
\textsuperscript{*}Equal contribution
}

\begin{document}

\maketitle

\begin{abstract}
Context compression aims to shorten long context inputs with minimal information loss for LLM inference acceleration. 
While existing methods have shown promise, they typically rely on complex compression modules or compression-specific training, leaving the intrinsic capabilities of LLMs underexplored.
In contrast, this work reveals that a thinking model itself can naturally compress long contexts by organizing task-relevant information. 
We thus derive \textit{Thinking as Compression} (TaC), a new compression paradigm that treats thinking itself as compressed context. 
{Without relying on specific dedicated compressor}, TaC directly prompts the thinking model to generate thinking traces as the shortened context, already outperforming most representative compression methods.
Further, given that raw thinking output may struggle with budget control and shortcut behaviors, we introduce \textit{Thinking as Compression Constrained} (TaC-C), leveraging a simple reward-driven optimization framework to elicit intrinsic thinking as compact and controllable compressed context.
Experiments across four long-context QA benchmarks demonstrate that TaC-C consistently outperforms existing baselines. At 4x and 8x compression ratios, it surpasses the strongest competitor by 17.4\% and 23.4\% in average F1, and by 15.7\% and 21.7\% in average Exact Match Score (EM), respectively.

\end{abstract}

\section{Introduction}\label{sec:intro}

Large Language Models (LLMs) are increasingly used in Retrieval-Augmented Generation (RAG)~\cite{yang2018hotpotqa, 2024EXIT}, long-document understanding~\cite{bai2024longbench, li2024long}, and long-horizon agentic workflows~\cite{jin2025search, li2025search, feng2025group}. They must process ever longer input contexts to provide richer evidence, even though such contexts incur substantial inference overhead and introduce redundant or irrelevant information~\cite{tang2025gmsa,dai2025pretraining, tang2026comi}. Context compression is therefore essential for efficient and reliable long-context inference, aiming to retain task-critical information within a much shorter input.

Canonical context compression methods typically fall into two representative paradigms: task-agnostic and task-aware compression. As shown in Figure~1(a), task-agnostic methods shorten the context without considering the downstream query, either by pruning tokens~\cite{pan2024llmlingua} or by encoding the input into compact implicit representations, such as soft tokens~\cite{dai2025pretraining, ge2023context, tang2025gmsa} or model states~\cite{li2024snapkv, zhang2024long}. They are broadly applicable, but may miss details that matter only for a specific query. In contrast, as illustrated in Figure~1(b), task-aware methods condition compression on the query and explicitly estimate query-context relevance to decide which context segments to retain~\cite{tang2026comi, jiang2024longllmlingua, 2024EXIT, 2025Provence, tang2026read}. However, they still mostly select or preserve existing fragments, rather than actively gathering, connecting, and reorganizing scattered information. Along these lines, existing methods largely treat compression as a separate mechanism designed specifically for shortening context, leaving the intrinsic compression ability of LLMs themselves underexplored.

In this work, we explore context compression from the perspective of model intrinsic capabilities. 
Given that thinking is fundamentally a process of distilling and compressing information~\cite{wei2022chain,uniscompression}, we propose \textbf{\textit{Thinking as Compression}} ({\color[RGB]{0, 81, 186}{\textit{\textbf{TaC}}}}), a new compression paradigm that treats thinking traces itself as compressed context.
As illustrated in Figure~1(c), thinking itself is potentially to compresses long contexts by focusing on key information, skipping redundancies, revisiting important evidence, and linking scattered facts into a compact trace for downstream generation. 
To validate this paradigm, we conduct a pilot study in Section~\ref{sec:pilot}, demonstrating that raw thinking traces can indeed serve as effective compressed contexts, already outperforming many representative compression methods.



\begin{figure}[!t]
  \centering
  \includegraphics[width=0.98\columnwidth]{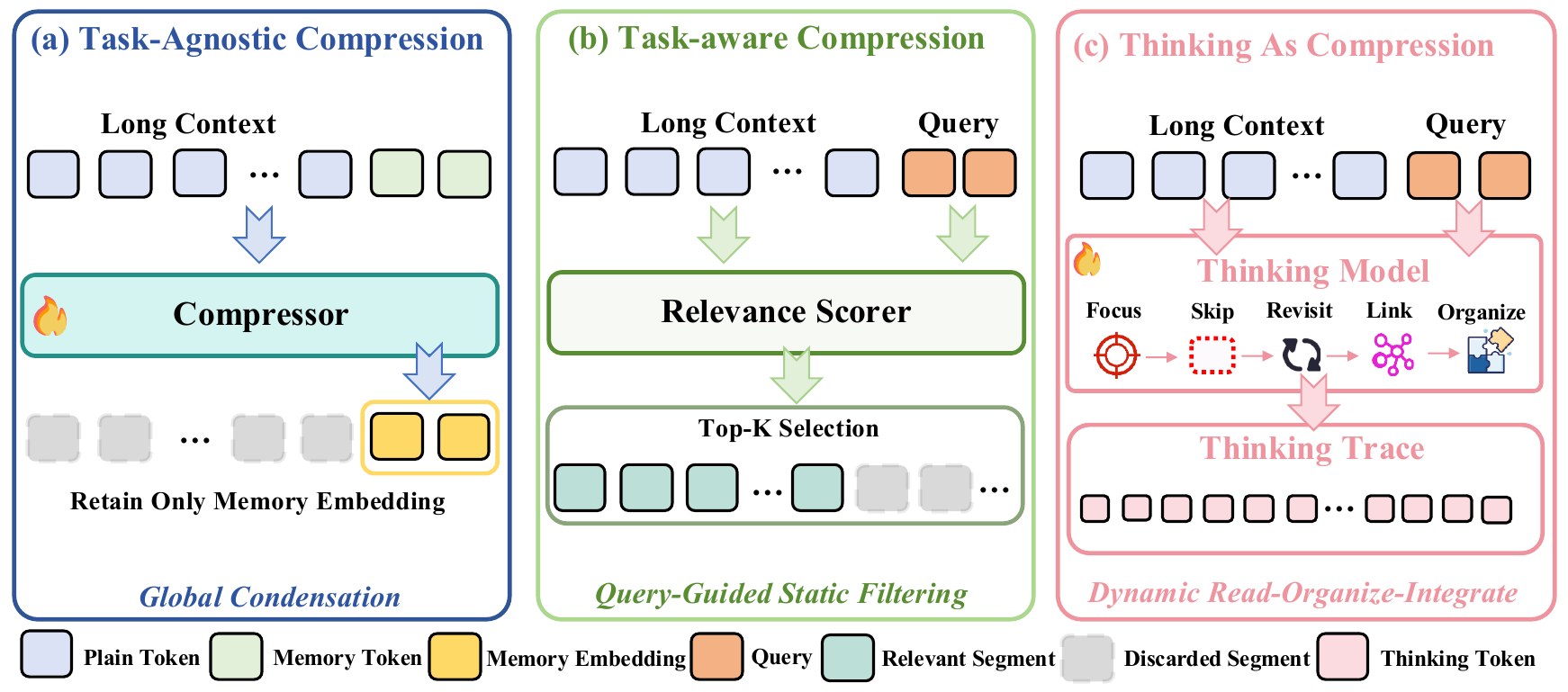}
  \caption{Comparison of context compression paradigms.
(a) Task-agnostic compression globally condenses long context into query-independent memory representations.
(b) Task-aware compression uses the query to select relevant context segments before decoding.
(c) Thinking As Compression leverages the model’s thinking ability to organize task-relevant context into a compact trace.}
  \label{fig:intro}

\end{figure}

However, the pilot study also shows that raw thinking traces, despite their usefulness, are not reliably aligned with the target budget or downstream generation needs, motivating explicit optimization to make them compact, controllable, and utility-preserving. To this end, we further propose \textbf{\textit{Thinking as Compression Constrained}} ({\color[RGB]{0, 81, 186}{\textit{\textbf{TaC-C}}}}), a simple reward-driven optimization approach that encourages the thinking model to produce compact and useful traces under a target budget. Specifically, \methodname{} straightforwardly combines utility and budget-control rewards with an anti-hack constraint, showing that a simple optimization scheme suffices to elicit the intrinsic organization ability of model -- yielding a more effective and controllable context compression mechanism. To summarize, our contributions are as follows:
\begin{itemize}
    \item We present \textit{Thinking as Compression} (\textit{TaC}), a new compression paradigm that views thinking traces as query-conditioned, dynamically organized representations of long contexts.

    \item We introduce a simple reward-driven optimization framework, \textit{Thinking as Compression Constrained} (\textit{TaC-C}), that elicits compact and useful reasoning traces for downstream use, while discouraging shortcut-oriented compression through an anti-hack constraint.

    \item We evaluate \methodname{} on long-context QA benchmarks and show that it consistently outperforms existing compression methods under 4$\times$ and 8$\times$ compression constraints, with strong transferability across scenarios and downstream models.
\end{itemize}




\section{TaC: Thinking as Compression}


In this section, we formalize \textit{Thinking as Compression} and conduct a pilot study to examine its potential as a context compression paradigm and the key factors that affect it.

\subsection{Paradigm Formulation}

Given a query $q$ and a long context $C=(c_1,\ldots,c_n)$ with length $L$.
In \textit{Thinking as Compression}, the model first produces a compact thinking trace $o=(o_1,\ldots,o_m)$ from the full input, and this trace is then treated as the compressed context for downstream generation:
\begin{equation}
    o \sim p_\theta(\cdot \mid q, C, \mathcal{B}), \quad
    \hat{y} \sim p_\theta(\cdot \mid q, o), \quad
    |o| \le \mathcal{B} .
\end{equation}
where $\mathcal{B} \ll L$ is the prescribed compression budget. Here, $o$ serves as the compressed context $\tilde{C}$, but it is generated through the model's thinking process rather than a separate compression operation.

\subsection{Pilot Study: How Thinking Impacts Context Compression?}
\begin{wrapfigure}{r}{0.42\columnwidth}
    \vspace{-0.5em}
    \centering
    \includegraphics[
        width=\linewidth,
        trim=0 0 0 0,
        clip
    ]{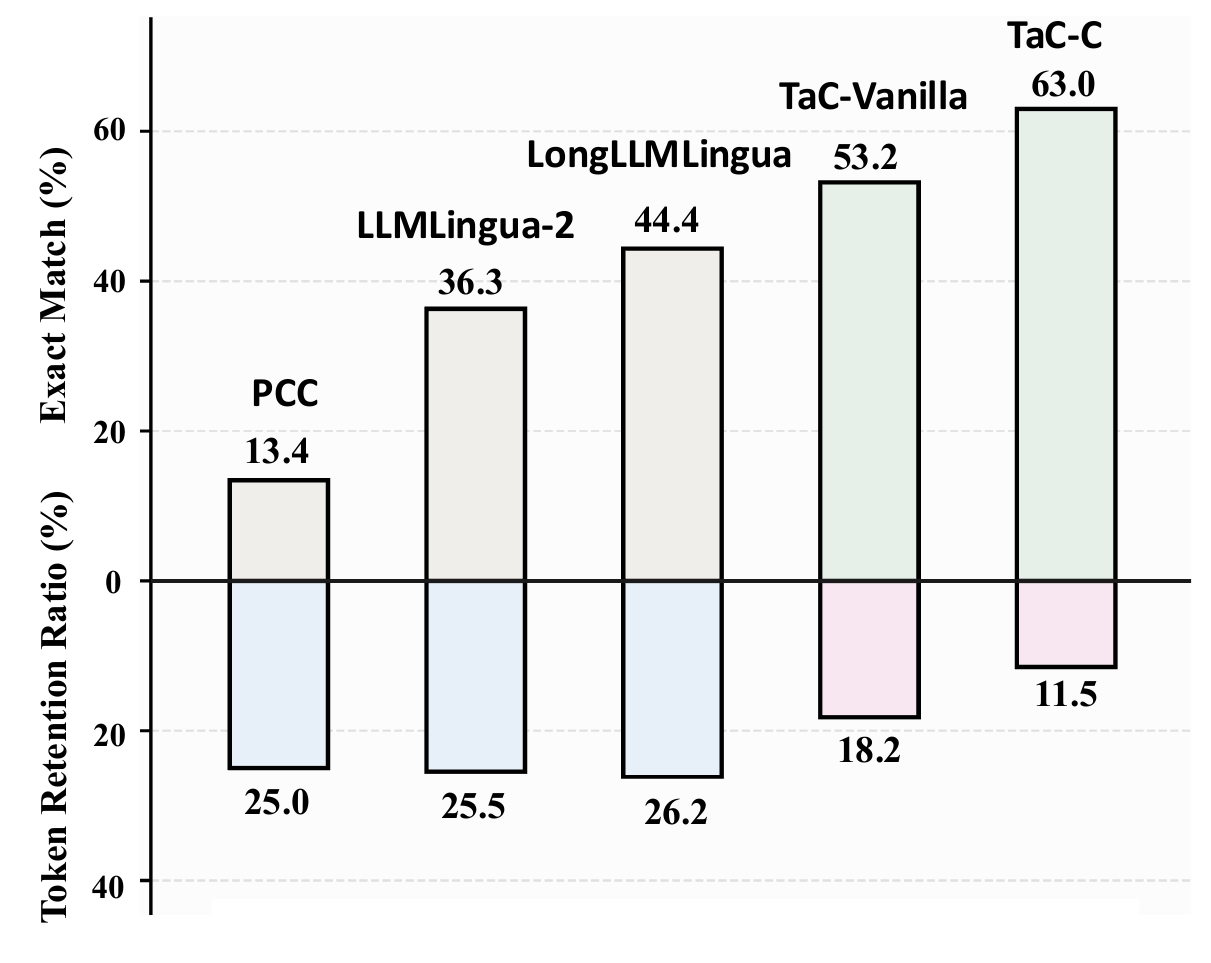}
    \vspace{-0.7em}
    \caption{Pilot study of {Thinking as Compression}. \textit{Token retention ratio} (compressed length / original length) measures the degree of compression.}
    \label{fig:pilot}
    \vspace{-0.8em}
\end{wrapfigure}
\label{sec:pilot}
To examine whether raw thinking traces can serve as compressed contexts, we instantiate a prompt-only variant of \textit{Thinking as Compression}, denoted as TaC-Vanilla. Given $(q,C)$ and a target budget, TaC-Vanilla prompts Qwen3-8B to generate a thinking trace $o$, which serves as the only context for downstream generation.
We evaluate it along two dimensions: compression quality, measured at the target 4$\times$ compression level, and budget controllability, measured by the actual compression ratio achieved.


As shown in Figure~\ref{fig:pilot}, the pilot study leads to two main observations:
\textit{i) Raw thinking traces show strong potential as reusable compressed contexts.}
Under comparable compression ratios, TaC-Vanilla substantially outperforms representative baselines in EM and F1, indicating that the model's intrinsic compression and organization ability can already produce useful natural-language traces for downstream generation.
\textit{ii) Prompt-only thinking lacks reliable budget controllability.}
On the full evaluation set, TaC-Vanilla achieves non-trivial compression but fails to consistently reach the target 4$\times$ budget. This indicates that prompt-only generation can induce compression behavior, but does not provide reliable control over the resulting trace length.

The pilot study shows that raw thinking traces provide a promising starting point for context compression, but they still need to be optimized for downstream usefulness, budget controllability, and reusable-context alignment. These findings motivate a lightweight optimization strategy that explicitly shapes thinking traces into useful, compact, and reliable compressed contexts, which we develop in the next section.

\section{TaC-C: Thinking as Compression Constrained}

\begin{figure}[!t]
  \centering
  \includegraphics[width=\columnwidth]{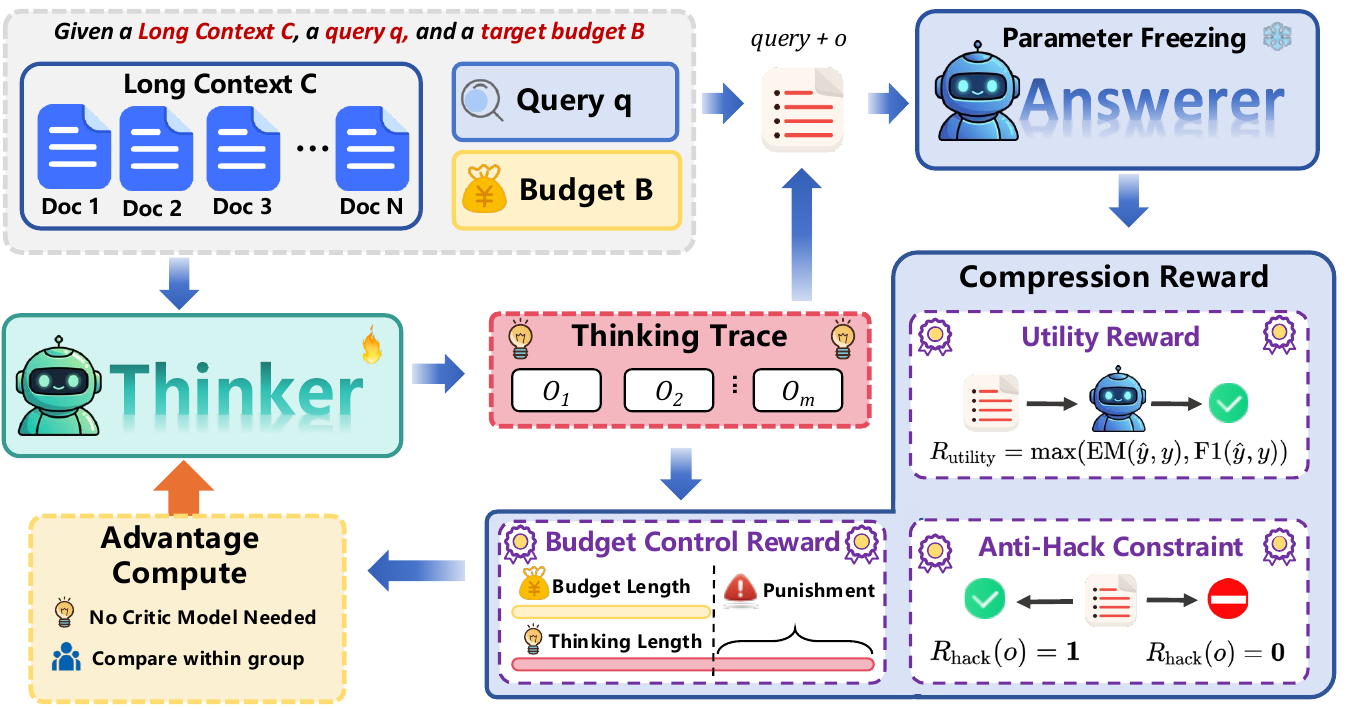}
  \caption{TaC-C framework. Given a long context C, query q, and budget B, the Thinker generates multiple compressed thinking traces, which are scored by utility, budget control, and anti-hacking rewards. GRPO then uses group rewards, advantages, and KL regularization to update the Thinker, enabling it to produce concise, task-relevant traces for the Answerer.}
  \label{fig:method}
\end{figure}

Building on the pilot study in section \ref{sec:pilot}, we propose \methodname{}, a RL-based framework for turning a reasoning model into a budget-aware context compressor. 
The goal is to elicit thinking traces that preserve downstream utility, satisfy the target budget, and remain aligned with the role of compressed context. As illustrated in Figure~\ref{fig:method}, \methodname{} adopts a decoupled Thinker--Answerer pipeline: the Thinker generates a trace $o$ from $(q,C, \mathcal{B})$, and the Answerer produces the final prediction from $(q, o)$ without accessing the original context.

We optimize the Thinker Model using Group Relative Policy Optimization (GRPO)~\cite{shao2024deepseekmath, guo2025deepseek} with reward mechanism based on pilot study. 
Specifically, the reward design directly follows the two factors identified in the pilot study. 
Trace informativeness is captured by downstream utility and budget controllability is enforced through a budget reward. 
Moreover, existing studies have revealed that reasoning answer propagation within LLMs may lead to mutual interference and biased path generation \cite{lin2025implicit, jiang2019avoiding}. 
To mitigate such undesirable cheating behavior, we further introduce a hacking-aware reward mechanism. 
We detail these components below.

\fakeparagraph{Format and Utility Reward}
We first introduce a binary format reward to ensure that the Thinker Model outputs a valid thinking trace. Specifically, the format reward is $1$ if the response contains a non-empty closed \texttt{<think>...</think>} block, and $0$ otherwise. This provides the minimal structural basis for evaluating all subsequent reward components.
On top of this, we define a utility reward under a decoupled pipeline with a trainable \textit{Thinker Model} and a frozen \textit{Answer Model}. Given $(q,C, \mathcal{B})$, the Thinker Model first generates a thinking trace $o \sim \pi_\theta(\cdot \mid q,C, \mathcal{B})$. The Answer Model, without access to the original long context, then relies solely on $q$ and $o$ for downstream prediction:
\begin{equation}
    \hat{y} = M_{\mathrm{ans}}(q, o), \qquad
    R_{\mathrm{utility}} = \max\!\left(\mathrm{EM}(\hat{y}, y), \mathrm{F1}(\hat{y}, y)\right).
\end{equation}
where $y$ denotes the ground-truth target.
This design is intended to elicit the Thinker Model's intrinsic reasoning ability for information compression. Using the Answer Model as a surrogate evaluator, the Thinker Model is rewarded according to whether its trace preserves task-critical evidence while filtering noisy content. We define the utility reward as $\max(\mathrm{EM},\mathrm{F1}) \in [0,1]$, where Exact Match(EM)~\cite{yang2018hotpotqa} captures precise task-level correctness and token-level F1~\cite{yang2018hotpotqa} provides a denser estimate of partial utility. This yields a more informative signal for learning compact and useful thinking traces.

\fakeparagraph{Budget Control Reward}
Although longer thinking traces can improve downstream performance, much of the intermediate reasoning is redundant for compression. In our setting, the thinking trace should act as a compact, high-density carrier of task-critical information rather than a verbose record of unconstrained reasoning. We therefore introduce a budget-aware reward to encourage the Thinker Model to concentrate useful evidence and express it succinctly within a limited budget.

Concretely, let $|o|$ denote the token length of the generated thinking trace and $\mathcal{B}$ the target budget. Before being passed to the Answer Model, the trace is truncated to at most $\mathcal{B}$ tokens, making any content beyond the budget invisible downstream. This forces the Thinker Model to organize task-critical evidence and useful relations within the budget.
To further stabilize training, we introduce a soft budget gate with a small tolerance window $\gamma$:
\begin{equation}
R_{\mathrm{budget}}(o;B)=
\begin{cases}
0, & |o|=0 \ \text{or}\ |o|\ge \mathcal{B}(1+\gamma),\\
1, & 0<|o|\le \mathcal{B},\\
1-\dfrac{|o|-\mathcal{B}}{\gamma \mathcal{B}}, & \mathcal{B}<|o|<\mathcal{B}(1+\gamma).
\end{cases}
\end{equation}
Under this design, in-budget traces receive full reward, mildly overlong traces are penalized smoothly, and excessively long traces receive zero reward. The tolerance window stabilizes optimization while still encouraging useful information to stay within budget.

\fakeparagraph{Anti-Hacking Constraints}
In our framework, the thinking trace is intended to function as compressed working memory rather than a shortcut to the final outcome. When the model exploits the reward by bypassing informative intermediate compression, the trace no longer serves as meaningful context and instead collapses into reward hacking behavior.
To preserve the intended role of the thinking trace, we introduce a rule-based anti-hacking gate:
\begin{equation}
R_{\mathrm{hack}}(o)=
\begin{cases}
0, & \text{if }\mathrm{Hack}(o)=1,\\
1, & \text{otherwise},
\end{cases}
\end{equation}
where $\mathrm{Hack}(o)$ indicates whether the generated trace exhibits reward-hacking patterns, such as direct answer disclosure. Once detected, the sample receives zero reward. This hard constraint discourages shortcut optimization and encourages the trace to remain an informative compressed representation.

\fakeparagraph{Optimization Objective}
Combining the above components, we define the overall reward for a generated thinking trace $\tau$ as
\begin{equation}
R(\tau)
=
R_{\mathrm{hack}}(o)
\left(
\lambda_{\mathrm{fmt}} R_{\mathrm{format}}(o)
+
\lambda_{\mathrm{utility}} R_{\mathrm{utility}}(o)\, R_{\mathrm{budget}}(o)
\right),
\end{equation}\label{eq:optimization_objective}
where $R_{\mathrm{format}}$ is the binary format reward, $R_{\mathrm{utility}} \in [0,1]$ measures downstream utility, $R_{\mathrm{budget}} \in [0,1]$ enforces budget adherence, and $R_{\mathrm{hack}} \in \{0,1\}$ is the anti-hacking gate. We optimize the Thinker Model by maximizing this reward with GRPO.
The objective encourages compact, length-controllable, and interpretable traces that preserve task-critical evidence and can be directly used by downstream models without architectural changes.

\section{Experiments}\label{sec:exp}
\begin{table*}[!ht]
\caption{Experimental results on four QA benchmarks. \textbf{Bold} denotes the best results. \textbf{F1} and \textbf{EM} denote F1 score and Exact Match, respectively. \textbf{Closed-book} uses only the question as input, whereas \textbf{Original Prompt} leverages the full context.}
\label{tab:main_results}
\centering
\small
\setlength{\tabcolsep}{5pt}
\renewcommand{\arraystretch}{0.92}
\resizebox{0.98\textwidth}{!}{%
\begin{tabular}{lcccccccccc}
\toprule
\multirow{2}{*}{\textbf{Methods}}
& \multicolumn{2}{c}{\textbf{NaturalQA}}
& \multicolumn{2}{c}{\textbf{2WikiMQA}}
& \multicolumn{2}{c}{\textbf{HotpotQA}}
& \multicolumn{2}{c}{\textbf{MuSiQue}}
& \multicolumn{2}{c}{\textbf{AVG}} \\
\cmidrule(lr){2-3}
\cmidrule(lr){4-5}
\cmidrule(lr){6-7}
\cmidrule(lr){8-9}
\cmidrule(lr){10-11}
& \textbf{F1} & \textbf{EM}
& \textbf{F1} & \textbf{EM}
& \textbf{F1} & \textbf{EM}
& \textbf{F1} & \textbf{EM}
& \textbf{F1} & \textbf{EM} \\
\midrule

\multicolumn{11}{c}{\textbf{LLaMA-3.1-8B-Instruct}} \\
\midrule
Closed-book       & 36.51 & 23.77 & 21.15 & 14.60 & 26.26 & 17.89 & 7.43  & 2.48  & 22.84 & 14.69 \\
Original Prompt   & 56.04 & 43.28 & 54.03 & 44.83 & 70.08 & 55.30 & 42.27 & 30.62 & 55.61 & 43.51 \\
EXIT              & 38.32 & 25.69 & 23.45 & 14.54 & 40.29 & 29.22 & 9.11  & 4.05  & 27.79 & 18.38 \\
Provence          & 51.27 & 38.57 & 43.08 & 32.51 & 59.79 & 44.82 & 26.65 & 17.87 & 45.20 & 33.44 \\
\midrule

\multicolumn{11}{c}{\textit{4x Compression Constraint}} \\
\midrule
LLMLingua-2-large & 46.66 & 33.15 & 37.07 & 30.38 & 50.89 & 36.95 & 29.95 & 20.69 & 41.14 & 30.29 \\
LongLLMLingua     & 58.25 & 45.61 & 39.32 & 33.09 & 57.41 & 41.55 & 35.22 & 23.83 & 47.55 & 36.02 \\
ICAE              & 50.35 & 38.49 & 35.59 & 28.98 & 37.95 & 26.97 & 18.73 & 9.02  & 35.66 & 25.87 \\
Activation Beacon & 58.97 & 47.04 & 44.20 & 36.69 & 57.44 & 43.78 & 51.27 & 38.19 & 52.97 & 41.43 \\
PCC               & 42.11 & 29.72 & 47.03 & 38.85 & 45.57 & 31.10 & 25.44 & 14.98 & 40.04 & 28.66 \\
\midrule
\rowcolor{rowblue}
\textbf{\methodname{}}     & \textbf{67.11} & \textbf{53.11} & \textbf{77.70} & \textbf{69.34} & \textbf{75.70} & \textbf{59.44} & \textbf{59.85} & \textbf{46.67} & \textbf{70.09} & \textbf{57.14} \\
\midrule

\multicolumn{11}{c}{\textit{8x Compression Constraint}} \\
\midrule
LLMLingua-2-large & 36.36 & 24.14 & 27.23 & 22.84 & 37.60 & 26.28 & 18.42 & 10.53 & 29.90 & 20.95 \\
LongLLMLingua     & 51.97 & 38.27 & 33.47 & 28.84 & 50.21 & 36.27 & 26.60 & 16.18 & 40.56 & 29.89 \\
ICAE              & 46.99 & 35.89 & 36.33 & 30.39 & 36.79 & 26.43 & 16.06 & 7.28  & 34.04 & 25.00 \\
Activation Beacon & 55.09 & 41.21 & 42.98 & 35.53 & 48.57 & 36.29 & 44.85 & 31.40 & 47.87 & 36.11 \\
PCC               & 49.59 & 37.02 & 38.10 & 31.55 & 43.19 & 30.02 & 25.66 & 16.59 & 39.14 & 28.80 \\
\midrule
\rowcolor{rowblue}
\textbf{\methodname{}}     & \textbf{69.03} & \textbf{55.74} & \textbf{77.39} & \textbf{69.29} & \textbf{75.89} & \textbf{59.52} & \textbf{57.78} & \textbf{44.19} & \textbf{70.02} & \textbf{57.19} \\
\midrule

\multicolumn{11}{c}{\textbf{Qwen3-8B}} \\
\midrule
Closed-book       & 22.86 & 11.17 & 27.58 & 22.15 & 25.64 & 16.27 & 10.75 & 2.98  & 21.71 & 13.14 \\
Original Prompt   & 60.44 & 46.40 & 48.45 & 39.33 & 67.68 & 53.06 & 34.69 & 23.13 & 52.85 & 40.48 \\
EXIT              & 31.20 & 21.02 & 32.33 & 27.93 & 41.68 & 30.70 & 13.52 & 5.05  & 29.68 & 21.18 \\
Provence          & 48.19 & 33.60 & 47.42 & 40.94 & 61.31 & 47.47 & 29.39 & 18.62 & 46.58 & 35.16 \\
\midrule

\multicolumn{11}{c}{\textit{4x Compression Constraint}} \\
\midrule
LLMLingua-2-large & 43.32 & 28.96 & 35.60 & 29.76 & 48.67 & 34.99 & 25.88 & 15.97 & 38.37 & 27.42 \\
LongLLMLingua     & \textbf{57.30} & \textbf{43.50} & 39.34 & 32.78 & 56.80 & 41.44 & 36.63 & 24.37 & 47.52 & 35.52 \\
ICAE              & 32.43 & 21.69 & 28.64 & 23.06 & 32.37 & 22.08 & 12.15 & 4.22  & 26.40 & 17.76 \\
PCC               & 44.06 & 25.46 & 25.93 & 19.47 & 33.04 & 19.78 & 17.05 & 9.85  & 30.02 & 18.64 \\
\midrule
\rowcolor{rowblue}
\textbf{\methodname{}}     & 52.52 & 36.53 & \textbf{78.47} & \textbf{68.83} & \textbf{74.28} & \textbf{57.71} & \textbf{55.19} & \textbf{41.62} & \textbf{65.12} & \textbf{51.17} \\
\midrule

\multicolumn{11}{c}{\textit{8x Compression Constraint}} \\
\midrule
LLMLingua-2-large & 33.22 & 20.49 & 29.85 & 26.33 & 36.35 & 25.20 & 17.63 & 9.64  & 29.26 & 20.42 \\
LongLLMLingua     & 49.61 & 35.33 & 34.97 & 29.80 & 49.60 & 35.11 & 29.02 & 17.34 & 40.80 & 29.40 \\
ICAE              & 35.67 & 23.47 & 30.91 & 25.36 & 32.85 & 22.27 & 12.21 & 4.80  & 27.91 & 18.98 \\
PCC               & 35.84 & 19.44 & 25.16 & 19.20 & 23.91 & 12.43 & 12.57 & 5.17  & 24.37 & 14.06 \\
\midrule
\rowcolor{rowblue}
\textbf{\methodname{}}     & \textbf{53.08} & \textbf{37.63} & \textbf{78.52} & \textbf{69.17} & \textbf{74.62} & \textbf{58.08} & \textbf{55.45} & \textbf{42.28} & \textbf{65.42} & \textbf{51.79} \\
\bottomrule
\end{tabular}%
}
\vspace{-0.2cm}
\end{table*}

\subsection{Settings}
\fakeparagraph{Training Setup}
We instantiate \methodname{} on two open-source backbones with strong reasoning capability, \texttt{Llama-3.1-8B-Instruct}~\cite{grattafiori2024llama} and \texttt{Qwen3-8B}~\cite{2025Qwen3}. We conduct experiments on four long-context question answering benchmarks: NaturalQuestions~\cite{kwiatkowski2019natural}, 2WikiMQA~\cite{ho2020constructing}, HotpotQA~\cite{yang2018hotpotqa}, and MuSiQue~\cite{trivedi2022musique}. To construct the training set, we randomly sample 3,000 instances from these datasets. During training, we adopt a mixed-budget setting in which each instance is assigned a compression ratio sampled from $\{4,8\}$. This allows a single \methodname{} model to generalize across different compression budgets at downstream inference time. Additional implementation and dataset details are provided in \appref{app:implement} and \appref{app:datasets}.
Following~\cite{2024EXIT}, we adopt Exact Match (EM) and F1~\cite{yang2018hotpotqa} as evaluation metrics. EM evaluates exact answer correctness, while F1 captures token-level overlap and partial matching, jointly assessing the quality and effectiveness of compressed contexts for downstream QA tasks.

\fakeparagraph{Baselines}
We conduct a comprehensive comparison with context compression and KV cache compression methods, covering three categories: i) hard prompt compression, including LLMLingua-2-large~\cite{pan2024llmlingua} and LongLLMLingua~\cite{jiang2024longllmlingua}; ii) soft prompt compression, including ICAE~\cite{ge2023context} and PCC~\cite{dai2025pretraining}; and iii) KV cache compression, represented by Activation Beacon~\cite{zhang2024long}. In addition, we include compression methods designed for downstream RAG tasks, such as EXIT~\cite{2024EXIT} and Provence~\cite{2025Provence}.

\subsection{Main Results}


As shown in Table~\ref{tab:main_results}, \methodname{} achieves the best overall performance across four long-context QA benchmarks under both 4$\times$ and 8$\times$ compression constraints. We highlight four key findings from experimental results: \textbf{\textit{i) Superior performance.}} \methodname{} consistently outperforms existing compression baselines in both F1 and EM, demonstrating that explicitly optimized thinking traces can serve as high-quality compressed contexts that preserve task-critical evidence and reorganize dispersed information for downstream generation. At 4x and 8x compression constraints, \methodname{} improves over the strongest competing baseline by 17.4\% and 23.4\% in average F1, respectively, and by 15.7\% and 21.7\% in average EM, respectively. Notably, \methodname{} even surpasses the full-context \textit{Original Prompt} setting in many cases, suggesting that redundant or distracting context can hurt generation, while effective compression can improve it by concentrating the input on query-relevant information. \textbf{\textit{ii) Advantage over query-based static filtering.}} 
Compared with query-based compressors such as EXIT, Provence, and LongLLMLingua, which mainly rely on static relevance or perplexity-based filtering, \methodname{} achieves stronger results by eliciting the reasoning model to actively integrate scattered evidence and reshape it into a task-oriented context. \textbf{\textit{iii) General applicability.}} the improvements are consistent across both LLaMA-3.1-8B-Instruct and Qwen3-8B. This demonstrates that \methodname{} is not model-specific, but provides a general framework for eliciting compression behavior from reasoning models and transferring the resulting traces to different downstream answer models. \textbf{\textit{iv) Robustness under stronger compression.}} \methodname{} remains robust under tighter compression. While most baselines degrade from 4$\times$ to 8$\times$ due to greater information loss, \methodname{} remains robust under tighter compression, maintaining stable performance from 4$\times$ to 8$\times$ while most baselines degrade. Its budget-aware reward encourages concise but flexible traces, enabling the Thinker to preserve high-value evidence within limited budgets.

\subsection{Analysis of Thinking Traces}

As shown in Table~\ref{tab:thinking_trace_cases}, the thinking traces generated by our lightly trained model exhibit diverse behaviors for context compression, including information extraction, summarization, and association. These patterns correspond to the core abilities required by query-conditioned compression: extracting the exact evidence needed by a query, aggregating dispersed information into a concise summary, and dynamically organizing related entities or events into a coherent context. Unlike task-agnostic compression methods that mainly shorten the input globally, or query-based filtering methods that rely on static relevance scores, our method uses the model's own reasoning ability to organize information according to the specific query. This suggests that a reasoner model can naturally act as a context compressor, and with simple control and guidance, its thinking trace can be effectively aligned with the compression objective to produce compact, coherent, and query-relevant contexts.

\begin{table*}[t]
\centering
\small
\setlength{\tabcolsep}{4pt}
\renewcommand{\arraystretch}{1.08}
\caption{Examples of thinking traces that support task-relevant compression. 
The traces exhibit \textbf{Information Extraction}, 
\textbf{Summarization}, and 
\textbf{Association}.}

\resizebox{0.98\textwidth}{!}{%
\begin{tabularx}{\textwidth}{
    >{\raggedright\arraybackslash}p{1.9cm}
    >{\raggedright\arraybackslash}p{2.2cm}
    >{\raggedright\arraybackslash}X
    >{\raggedright\arraybackslash}p{2.3cm}
}
\toprule[1.1pt]
\textbf{Type} & \textbf{Query} & \textbf{Thinking Trace} & \textbf{Label} \\
\midrule

\textbf{Extraction} &
When did Gina launch an ad campaign for her store? &
\graytrace{...Looking through the dates provided, the first mention of Gina launching an ad campaign is on \textcolor{black}{\textbf{29 January 2023}}. In that conversation, Gina says, ``I just launched...} &
\textbf{29 January, 2023} \\

\midrule

\textbf{Summarization} &
Does Calvin wish to become more popular? &
\graytrace{...First, I'll scan the conversations for signs of Calvin's goals related to popularity. }
\textbf{The context highlights Calvin's music career, collaborations, performances, and his desire to expand his global brand and fanbase, suggesting that he wants to become more popular...} &
\textbf{Yes.} \\

\midrule

\textbf{Association} &
Who owns the record label of the Shake What God Gave Ya performer? &
\graytrace{... the album ``Shake What God Gave Ya'' is by James Otto, and it was released through \textcolor{black}{\textbf{Warner Bros. Nashville}} which is part of \textcolor{black}{\textbf{Warner Music Group}}...} &
\textbf{Warner Music Group} \\

\bottomrule[1.1pt]
\end{tabularx}
}
\label{tab:thinking_trace_cases}
\end{table*}




\subsection{Scaling Behavior of Thinker}

\begin{wrapfigure}[14]{r}{0.43\columnwidth}
    \vspace{-2em}
    \centering
    \includegraphics[width=\linewidth]{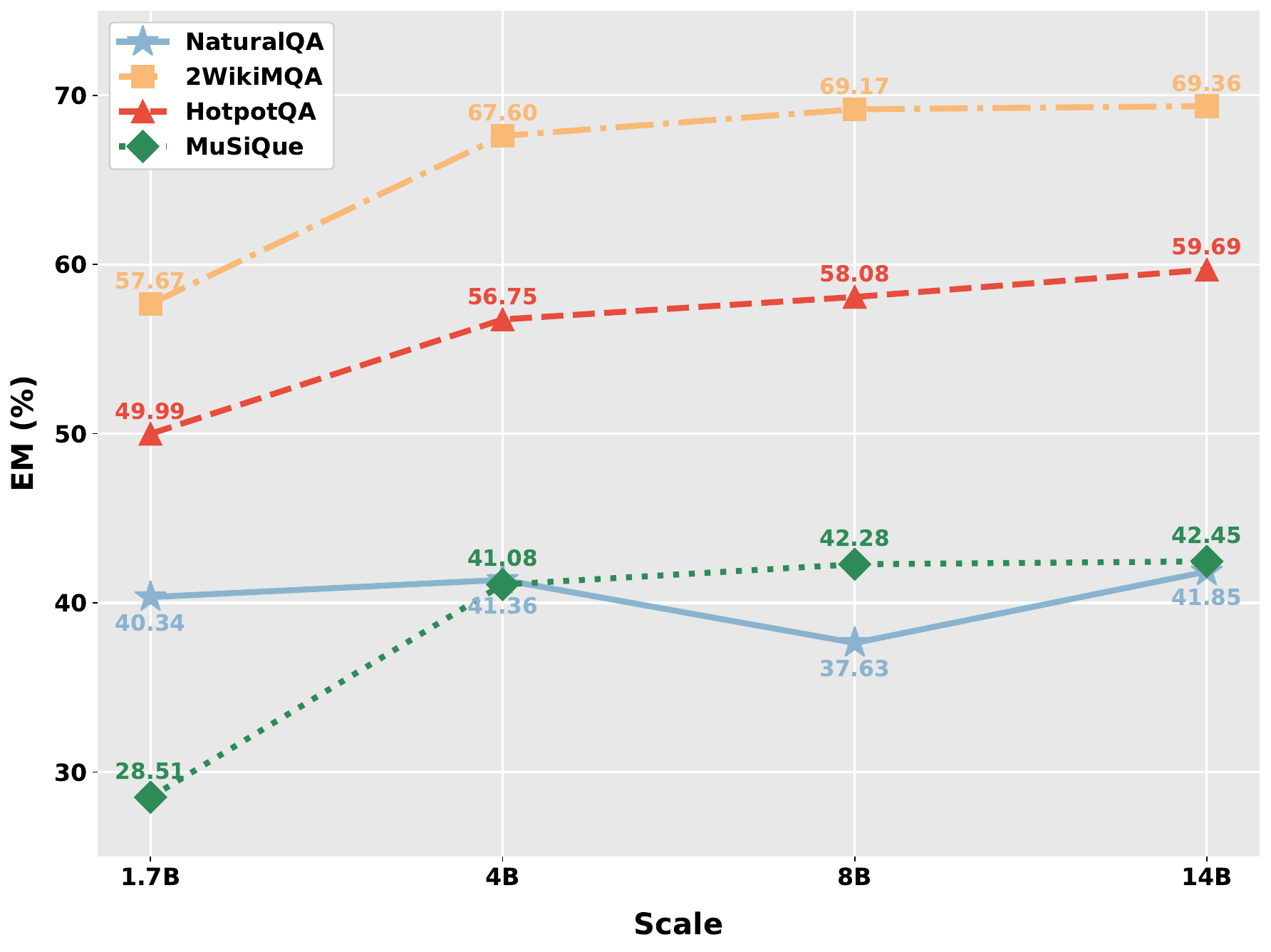}
    \vspace{-1.2em}
    \caption{ Exact Match (EM) scores on four datasets across different model scales.}
    \label{fig:thinker_scale}
    \vspace{-0.9em}
\end{wrapfigure}
To examine how the scale of the Thinker Model affects context compression, we fix Qwen3-8B as the Answer Model and train \methodname{} with Qwen3-family Thinkers of different scales, from 1.7B to 14B. Figure~\ref{fig:thinker_scale} reports dataset-wise F1 and EM scores across the four benchmarks.
The results show a clear improvement from 1.7B to 4B, suggesting that stronger Thinkers generate more informative compressed traces. 
Beyond 4B, however, the gains become relatively modest, indicating that the compression ability begins to saturate at moderate model scales. These trends are consistent across datasets: performance improves most noticeably from 1.7B to 4B and then largely saturates, suggesting that moderate-scale Thinkers already provide sufficient reasoning capacity for effective compression. This highlights the practicality of \methodname{}, where small reasoners act as efficient compressors without sacrificing downstream performance.


\begin{table*}[h]
\caption{
Evaluation of transferability across Answer Models under the 8$\times$ compression setting. 
}
\label{tab:generalization_answer_models}
\centering
\small
\setlength{\tabcolsep}{4.2pt}
\renewcommand{\arraystretch}{1.12}
\begin{tabular}{lcccccccccc}
\toprule
\multirow{2}{*}{\textbf{Methods}}
& \multicolumn{2}{c}{\textbf{NaturalQA}}
& \multicolumn{2}{c}{\textbf{2WikiMQA}}
& \multicolumn{2}{c}{\textbf{HotpotQA}}
& \multicolumn{2}{c}{\textbf{MuSiQue}}
& \multicolumn{2}{c}{\textbf{AVG.}} \\
\cmidrule(lr){2-3}
\cmidrule(lr){4-5}
\cmidrule(lr){6-7}
\cmidrule(lr){8-9}
\cmidrule(lr){10-11}
& \textbf{F1} & \textbf{EM}
& \textbf{F1} & \textbf{EM}
& \textbf{F1} & \textbf{EM}
& \textbf{F1} & \textbf{EM}
& \textbf{F1} & \textbf{EM} \\
\midrule

\multicolumn{11}{c}{\textbf{LLaMA-3.1-8B-Instruct}} \\
\midrule
LLMLingua-2-large
& 36.36 & 24.14 & 27.23 & 22.84 & 37.60 & 26.28 & 18.42 & 10.53 & 29.90 & 20.95 \\
LongLLMLingua
& 51.97 & \textbf{38.27} & 33.47 & 28.84 & 50.21 & 36.27 & 26.60 & 16.18 & 40.56 & 29.89 \\
\rowcolor{rowblue}
\textbf{\methodname{}}
& \textbf{53.49} & 37.51
& \textbf{76.76} & \textbf{66.32}
& \textbf{72.37} & \textbf{54.89}
& \textbf{54.29} & \textbf{39.51}
& \textbf{64.23} & \textbf{49.56} \\

\midrule
\multicolumn{11}{c}{\textbf{Gemma-4-26B-A4B}} \\
\midrule
LLMLingua-2-large
& 29.28 & 16.38 & 12.12 & 7.14 & 23.47 & 14.47 & 11.07 & 5.83 & 18.98 & 10.96 \\
LongLLMLingua
& 43.63 & 29.27 & 15.02 & 9.05 & 40.48 & 27.60 & 22.00 & 12.95 & 30.28 & 19.72 \\
\rowcolor{rowblue}
\textbf{\methodname{}}
& \textbf{55.94} & \textbf{38.31}
& \textbf{77.81} & \textbf{68.31}
& \textbf{75.73} & \textbf{58.69}
& \textbf{55.93} & \textbf{41.41}
& \textbf{66.35} & \textbf{51.68} \\

\midrule
\multicolumn{11}{c}{\textbf{Qwen3-30B-A3B}} \\
\midrule
LLMLingua-2-large
& 37.54 & 24.75 & 29.23 & 25.75 & 38.13 & 26.93 & 17.90 & 9.39 & 30.70 & 21.70 \\
LongLLMLingua
& 53.82 & 40.00 & 35.14 & 29.77 & 51.62 & 37.06 & 28.37 & 16.47 & 42.24 & 30.82 \\
\rowcolor{rowblue}
\textbf{\methodname{}}
& \textbf{56.40} & \textbf{41.39}
& \textbf{78.46} & \textbf{69.02}
& \textbf{74.85} & \textbf{58.72}
& \textbf{55.68} & \textbf{42.66}
& \textbf{66.35} & \textbf{52.95} \\

\bottomrule
\end{tabular}
\vspace{-0.4cm}
\end{table*}

\subsection{Applicability Across Answer Models}
In this part, We evaluate whether the thinking traces generated by \methodname{} can be directly reused by different downstream Answer Models. Under the 8$\times$ compression setting, we use Qwen3-8B as the Thinker Model to generate traces, and feed the same traces to LLaMA-3.1-8B-Instruct~\cite{grattafiori2024llama}, Gemma-4-26B-A4B~\cite{google_gemma4_26b_a4b_2026}, and Qwen3-30B-A3B~\cite{2025Qwen3} for answer generation. We compare against two hard prompt compression baselines, LLMLingua-2-large~\cite{pan2024llmlingua} and LongLLMLingua~\cite{jiang2024longllmlingua}.

As shown in \tabref{tab:generalization_answer_models}, \methodname{} achieves the best performance across all Answer Models and benchmarks. Since the same thinking traces are generated by a fixed Thinker and directly consumed by different Answer Models, these results indicate that \methodname{} produces compressed contexts that are not tied to a specific downstream model. Instead, the traces organize task-relevant evidence in an explicit and reusable form that different models can interpret and exploit. Hard prompt baselines show weaker cross-model transferability, while soft-token methods often require model-specific alignment. In contrast, \methodname{} produces natural-language thinking traces that plug into different downstream models without architectural changes.

\subsection{Generalization and Efficiency}

\begin{wraptable}{r}{0.55\textwidth}
\vspace{-1em}
\centering
\caption{Comparison on LoCoMo under 4$\times$ compression ratio. Latency is measured in milliseconds (ms). \textit{E2E}. denotes the end-to-end latency. \textit{Act}. is the actual compression ratio averaged across datasets.}
\label{tab:data_generalization}
\vspace{-0.4em}

\resizebox{0.55\textwidth}{!}{
\begin{tabular}{lccccc}
\toprule
\textbf{Method}  
& \textbf{Prefill}
& \textbf{Decode}
& \textbf{E2E}
& \textbf{F1} 
& \textbf{Act.(\%)} \\
\midrule
PCC & 239.5 & 423.4 & 662.9 & 14.17 & 25.14 \\
LLMLingua-2 & 241.9 & 449.2 & 691.1 & 31.53 & 21.11 \\
LongLLMLingua & 501.8 & 1504.6 & 2006.4 & \textbf{43.94} & 22.79 \\
\rowcolor{rowblue}
\textbf{\methodname{}} & \textbf{77.9} & \textbf{402.7} & \textbf{480.6} & 41.94 & \textbf{1.13} \\
\bottomrule
\end{tabular}
}
\vspace{-0.5em}
\end{wraptable}
To evaluate the robustness of our method under distribution shift, we further conduct experiments on LoCoMo~\cite{maharana2024evaluating}, a benchmark designed for long-term conversational memory. Unlike standard document-based QA datasets, LoCoMo consists of multi-session dialogues and requires models to retrieve, summarize, and reason over information distributed across long conversational histories.
As shown in Table~\ref{tab:data_generalization}, TaC-C generalizes effectively to long conversational memory scenarios and achieves performance comparable to LongLLMLingua. Importantly, TaC-C attains this performance while using a substantially lower actual compression ratio (Act.). Moreover, TaC-C significantly reduces both prefilling and decoding latency, leading to the lowest end-to-end inference time among all methods. These results show that our method generalizes across data distributions while achieving a superior efficiency–performance trade-off for long-context inference.

\subsection{Ablation Study}

\begin{wraptable}{r}{0.55\textwidth}
\vspace{-1.0em}
\centering
\caption{Ablation results under the 8$\times$ compression constraint. 
\textit{Act}. denotes the actual compression ratio averaged across datasets.}
\label{tab:ablation_study}
\vspace{-0.5em}
\resizebox{0.55\textwidth}{!}{
\begin{tabular}{lcccc}
\toprule
\textbf{Config.} 
& \textbf{EM(\%)} 
& \textbf{F1(\%)} 
& \textbf{Act.(\%)}
& \textbf{Hack(\%)}\\
\midrule
\rowcolor{rowblue}
\textbf{\methodname{}}
& 51.79
& 65.42
& 9.90
& 0.88 \\
\midrule
w/o $R_{\mathrm{utility}}$
& 30.25 
& 41.40 
& 6.60 
& 1.17 \\
w/o $R_{\mathrm{budget}}$
& 58.25 
& 71.78 
& 36.20 
& 1.95 \\
w/o $R_{\mathrm{hack}}$
& 61.72 
& 73.54 
& 10.30 
& 71.67 \\
\bottomrule
\end{tabular}
}
\vspace{-0.7em}
\end{wraptable}

Table~\ref{tab:ablation_study} shows that each reward component is necessary for shaping \methodname{} into a reliable context compressor.
Removing the utility reward leads to a clear performance collapse, indicating that query guidance alone is insufficient for learning effective active compression. Since a good compressed trace is defined by whether it can support the downstream Answerer without access to the original context, feedback from the Answerer is necessary to teach the Thinker what information should be preserved.
Removing the budget reward or the anti-hack gate improves accuracy, but sacrifices the intended compression behavior: the former produces overlong traces with an actual compression ratio above 36\%, while the latter raises the hack rate to more than 70\% by encouraging answer-like shortcuts. These results show that the full \methodname{} achieves a better balance between downstream utility, budget control, and reusable-context alignment.


\section{Related Work}


\fakeparagraph{Context Compression}
Long-context compression has been widely studied to reduce the computational cost of long-context inference and to make long inputs usable under limited context windows. Existing methods can be broadly divided into task-agnostic and task-aware approaches. Task-agnostic methods compress contexts without conditioning on the downstream query, typically preserving global semantic fidelity through token pruning~\cite{pan2024llmlingua, pan2024llmlingua}, latent memory representations~\cite{ge2023context, dai2025pretraining, feldman2025simple, tang2025gmsa}, summarization~\cite{xu2023recomp}, and KV-cache compression~\cite{li2024snapkv, zhang2024long}. While these methods improve efficiency and retain coarse-grained semantics, they may discard information that is not globally salient but is essential for a specific query. Task-aware methods alleviate this issue by conditioning compression on the query, typically preserving relevant content while filtering out less useful parts according to query-context relevance~\cite{jiang2024longllmlingua, tang2026comi, tang2026read, 2024EXIT, 2025Provence}. However, compression remains largely a static relevance-based step, often relying on specialized modules or compression-specific training.
In contrast, our work explores whether context compression can be achieved without a specialized compressor. We treat the reasoner itself as a query-conditioned compressor and elicit its intrinsic reasoning ability to adaptively condense long contexts, offering a simpler path toward effective context compression.

\fakeparagraph{Chain Of Thought}
Chain-of-Thought (CoT) reasoning~\cite{wei2022chain} has become a central mechanism for improving the reasoning ability of large language models. By externalizing intermediate steps, CoT provides additional test-time computation before the final response, enabling models to solve complex tasks more reliably~\cite{yao2022react}. Subsequent methods, including self-consistency~\cite{wang2022self}, process supervision~\cite{lightman2023let}, and reinforcement learning~\cite{guo2025deepseek, yu2025dapo, zheng2025group}, further improve the robustness and quality of these thinking trajectories. Recent advanced reasoners such as DeepSeek-R1~\cite{guo2025deepseek} and OpenAI's o3~\cite{el2025competitive} demonstrate that scaling such test-time thinking is a powerful route to stronger model capabilities. This motivates viewing thinking traces not only as rationales, but also as optimizable intermediate representations that shape downstream generation.


Beyond its role in eliciting stronger reasoning, CoT provides an operational view of how information is accumulated during generation. Recent information-theoretic analyses characterize intermediate reasoning steps through information gain, where each effective step increases the predictive certainty of the final outcome~\cite{2024An, wang2024analyzing, 2024Understanding}. Complementary evidence further shows that the global structure of long reasoning traces is a key factor in eliciting advanced reasoning behavior, suggesting that information gain is not accumulated independently at isolated steps, but organized through a coherent reasoning trajectory~\cite{2025LLMs}. From this perspective, a thinking trace can be viewed as a structured intermediate representation that progressively transforms dispersed contextual evidence into a more determined form for downstream generation. This view naturally connects CoT with context compression: given a long context containing redundancy and distractions, a query-conditioned thinking trace can serve as a compact carrier of accumulated information gain. 

\section{Conclusion}\label{sec:conclusion}

In this work, we explore context compression through the intrinsic thinking ability of modern reasoning LLMs. We propose \textit{Thinking as Compression} (TaC), which treats thinking traces themselves as compressed contexts, avoiding dedicated compression modules. While prompt-only TaC already shows strong potential, raw thinking traces may suffer from poor budget control and shortcut behaviors. To address this, we introduce \methodname{}, a simple reward-driven framework that optimizes thinking traces to be compact, controllable, and useful for downstream generation. Experiments show that \methodname{} consistently outperforms existing compression baselines under different compression ratios and produces reusable compressed contexts that transfer across downstream models.

\fakeparagraph{Limitations} Our study investigates thinking traces as compressed contexts for long-context. While promising, extending thinking-as-compression to broader scenarios—code understanding, tool-use histories, agent trajectories—remains under-explored. Moreover, although our method proves effective in many cases, its performance on extremely long context inputs remains an open challenge.

\bibliographystyle{abbrv}
\bibliography{citations}


\appendix

\section{Implementation Details.}\label{app:implement}
All reinforcement learning experiments are implemented with \texttt{VeRL}~\cite{sheng2024hybridflow} using a pure GRPO objective~\cite{shao2024deepseekmath, guo2025deepseek}, without training a separate critic model. Llama3.1-8B-Instruct~\cite{grattafiori2024llama} and Qwen3-8B~\cite{2025Qwen3} are trained under an identical configuration. For each input prompt, 8 responses are generated with \texttt{vLLM}~\cite{kwon2023efficient}, where the maximum prompt length and response length are set to 8192 and 2048 tokens, respectively. Policy optimization is performed with LoRA~\cite{hu2022lora}, using a rank of 128, a scaling factor of 256, and applying adapters to all linear layers. The actor learning rate is $1\times 10^{-5}$, the global training batch size is 128, the PPO mini-batch size is 64, and the per-GPU micro-batch size is 4. KL regularization to the reference policy is applied with a coefficient of 0.01, and each model is trained for one epoch. The reward adopts the formulation described in Eq.~\ref{eq:optimization_objective}, with $\lambda_{\mathrm{fmt}}=0.05$ and $\lambda_{\mathrm{utility}}=0.95$. All experiments are conducted on 8 NVIDIA H800 80GB GPUs.

\section{Datasets}\label{app:datasets}
\subsection{Details}
\fakeparagraph{NaturalQuestions} NaturalQuestions~\cite{kwiatkowski2019natural} is a large-scale open-domain question answering dataset built from real Google search queries and Wikipedia articles. Its questions and answers reflect authentic user information needs rather than manually authored prompts. Each instance provides both short-answer spans and long-answer passages, supporting evaluation of precise answer extraction and long-context comprehension. In our setting, each example contains 20 documents, including one ground-truth document and 19 distractors.

\fakeparagraph{2WikiMQA} 2WikiMQA~\cite{ho2020constructing} is a multi-hop question answering dataset constructed from Wikipedia. It requires models to combine evidence from multiple documents and capture the relations between facts to answer complex questions. Compared with HotpotQA, 2WikiMQA emphasizes more structured reasoning patterns, such as comparison, bridging, and inference over entity relations.

\fakeparagraph{HotpotQA} HotpotQA~\cite{yang2018hotpotqa} is a multi-hop question answering benchmark built on Wikipedia. Each question requires evidence from multiple articles, encouraging models to identify relevant entities, connect supporting facts across documents, and produce answers through cross-document reasoning rather than single-span extraction.

\fakeparagraph{MuSiQue} MuSiQue~\cite{trivedi2022musique} is a multi-hop question answering dataset constructed through the composition of connected single-hop questions. Its design explicitly reduces shortcut solutions by ensuring that one reasoning step depends on information obtained from another. The dataset contains 2--4 hop questions and emphasizes genuine connected reasoning across multiple pieces of evidence, making it a challenging benchmark for evaluating compositional multi-hop reasoning.

\subsection{Statistics}
Table~\ref{tab:dataset_context_stats} summarizes the context length statistics of the datasets used in our experiments. The combined benchmark contains 24,993 examples from four long-context question answering datasets. These statistics indicate that the evaluated instances generally involve long and information-dense contexts, making them well suited for assessing the effectiveness of context compression methods.

\begin{table}[h]
\caption{Statistics of context token lengths across datasets.}
\label{tab:dataset_context_stats}
\centering
\small
\begin{tabular}{lcccc}
\toprule
\textbf{Dataset} & \textbf{\# Samples} & \textbf{Avg. Tokens} & \textbf{Min. Tokens} & \textbf{Max. Tokens} \\
\midrule
NaturalQuestions  & 2,655  & 2,933 & 2,437 & 4,340 \\
2WikiMultihopQA   & 12,576 & 1,018 & 201   & 4,367 \\
HotpotQA          & 7,345  & 1,402 & 430   & 3,567 \\
MuSiQue           & 2,417  & 2,536 & 1,078 & 5,315 \\
\midrule
Total             & 24,993 & 1,481 & --    & --    \\
\bottomrule
\end{tabular}

\end{table}

\section{Prompt}

In this section, we present the instruction prompt designed to align the Thinker Model with the role of a context compressor. The prompt encourages the generated \texttt{<think>} trace to serve as a high-density carrier of compressed context, preserving task-relevant information in a concise and structured form. By specifying both the maximum \texttt{<think>} length and the compression ratio, the prompt further promotes budget-aware compression.

\begin{tcolorbox}[
    colback=gray!5,
    colframe=black,
    boxrule=0.8pt,
    arc=1.5mm,
    left=2mm,
    right=2mm,
    top=2mm,
    bottom=2mm,
    breakable
]
\noindent{\color{blue}\textbf{Prompt for Thinker Model Training}}

\vspace{0.6em}
\noindent
You are a query-conditioned context compressor. Given the context and current information need, write a high-quality \texttt{<think>} trace that preserves the context information a downstream model needs to locate and use relevant information.

\vspace{0.5em}
\noindent
\textbf{Task Guidelines:}

\begin{enumerate}[leftmargin=1.8em]
    \item \textbf{Information Selection}

    \begin{itemize}[leftmargin=2em]
        \item For information-dense context, extract and summarize information-need-relevant facts, evidence, entities, and relations.
        \item For example-based context, identify relevant examples and summarize reusable notes on the underlying pattern, format, criterion, or strategy.
        \item For trajectory-level context, preserve the current state, completed work, useful outputs, decisions, errors, failed attempts, and experience.
        \item Connect scattered information and reorganize it into a concise, task-useful structure.
    \end{itemize}

    \item \textbf{Think Requirements}

    \begin{itemize}[leftmargin=2em]
        \item Keep \texttt{<think>} concise, structured, context-grounded, and relevant to the information need.
        \item Prefer compact summaries, relations, reusable criteria, and state updates over context restatement.
        \item Use \texttt{<think>} as a compressed trace for downstream use, not as a final-answer field.
        \item Do not solve the information need or state a final answer.
    \end{itemize}

    \item \textbf{Output Format} Your response should be structured as follows:

    \vspace{0.3em}
    \noindent{\color{blue}\texttt{<think>}}

    \texttt{[High-quality compressed context trace for downstream use.]}

    \noindent{\color{blue}\texttt{</think>}}
\end{enumerate}

\vspace{0.4em}
\noindent
\textbf{Context:} \texttt{\{context\}}

\noindent
\textbf{Information Need:} \texttt{\{question\}}

\noindent
\textbf{Compression Budget:}
\begin{itemize}[leftmargin=2em]
    \item Maximum \texttt{<think>} length: \texttt{\{target\_think\_len\}} tokens.
    \item Compression ratio: \texttt{\{comp\_ratio\}}$\times$ from the original \texttt{\{context\_token\_len\}} context tokens.
\end{itemize}
\end{tcolorbox}

\section{RL Training Dynamics}

\begin{figure}[!h]
\caption{Training curves of the three reward components in \methodname{}: utility reward, budget control reward, and anti-hacking constraint. Shaded regions denote $\pm$1 standard deviation.}
  \label{fig:reward_curve}
  \centering
  \includegraphics[width=\columnwidth]{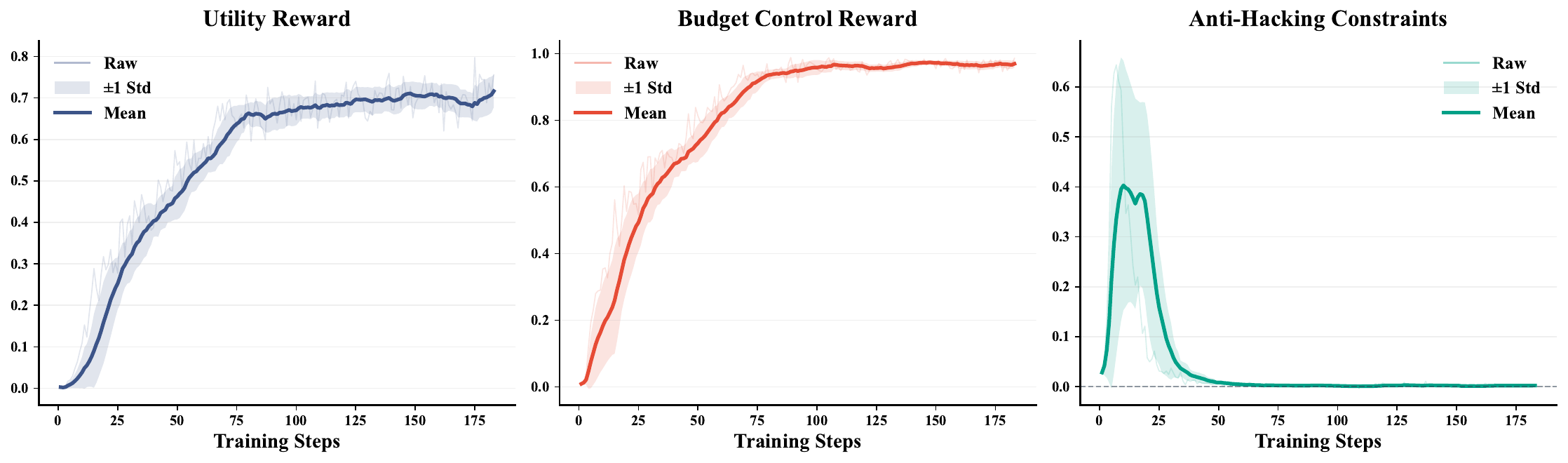}
\end{figure}

Figure~\ref{fig:reward_curve} shows how the reward design guides the reasoner’s thinking during training. As optimization proceeds, the utility reward increases, indicating that the thinking trace becomes more useful for downstream generation. Meanwhile, the budget reward rises and stabilizes, showing that the reasoner learns to express this useful information within the prescribed length constraint. The anti-hacking signal is mainly triggered in the early stage and then quickly diminishes, suggesting that shortcut-oriented thinking is suppressed during optimization. Together, these curves show that \methodname{} progressively guides the reasoner’s thinking from unconstrained problem solving toward compact information extraction and organization for context compression.


\section{Case Study Comparison}
In this case study, we examine how different compression methods preserve and organize information under a high compression ratio. LLMLingua-2 retains several important tokens, including the release years needed for comparison, but the compressed result becomes fragmented and lacks coherent organization. Although LongLLMLingua further incorporates the query to focus on relevant content, its metric-based filtering still removes some details that are necessary for answering the question. PCC keeps more surface-level contextual information after reconstruction, yet the recovered text is not fully faithful and introduces unsupported content, resulting in hallucination~\cite{2026When}. In contrast, TaC-C leverages the model's thinking process to analyze the query and identify the key evidence, thereby producing a faithful and well-structured compressed context for downstream use.

\vspace{-0.2cm}
\begin{tcolorbox}[
    breakable,
    title=\textbf{Case Study of Different Compression Methods},
    colback=gray!3,
    colframe=black!70,
    fonttitle=\bfseries,
    coltitle=white,
    colbacktitle=black!75,
    boxrule=0.7pt,
    arc=1.5pt,
    left=6pt,
    right=6pt,
    top=6pt,
    bottom=6pt
]

\textcolor{black}{\textbf{\large Context:}}

\vspace{0.4em}

\textbf{Document: 13 at a Table.}
\textit{13 at a Table} (also known as ``Tredici a tavola'') is a 2004 Italian comedy film written and directed by Enrico Oldoini. It was entered into the main competition at the 2005 Tokyo International Film Festival.

\vspace{0.45em}

\textbf{Document: S\~ao Paulo, Sociedade An\^onima.}
\textit{S\~ao Paulo, Sociedade An\^onima} is a 1965 Brazilian drama film written and directed by Luis S\'ergio Person. The film was selected as the Brazilian entry for the Best Foreign Language Film at the 38th Academy Awards, but was not accepted as a nominee.

\vspace{0.45em}

\textbf{Document: Trolleybuses in S\~ao Paulo.}
Trolleybuses in S\~ao Paulo provide a portion of the public transport service in Greater S\~ao Paulo, in the state of S\~ao Paulo, Brazil, with two independent trolleybus systems. The SPTrans system opened in 1949 and serves the city of S\~ao Paulo, while the Empresa Metropolitana de Transportes Urbanos de S\~ao Paulo (EMTU) system opened in 1988 and serves suburban areas to the southeast of the city proper.

\vspace{0.45em}

\textbf{Document: S\~ao Lu\'is.}
S\~ao Lu\'is or S\~ao Luiz may refer to several places in the Portuguese-speaking world.

\vspace{0.45em}

\textbf{Document: List of Governors of S\~ao Paulo.}
Governor of S\~ao Paulo is the position of the head of state and government of S\~ao Paulo, Brazil.

\vspace{0.45em}

\textbf{Document: Chafing dish.}
A chafing dish is a metal cooking or serving pan on a stand with an alcohol burner holding chafing fuel below it. It is used for cooking at table or as a food warmer for keeping dishes at a buffet warm.

\vspace{0.45em}

\textbf{Document: Copa S\~ao Paulo de Futebol J\'unior.}
The Copa S\~ao Paulo de Futebol J\'unior, also known as Copa S\~ao Paulo de Juniores and Copinha, is a cup competition played by Brazilian under-20 football teams.

\vspace{0.45em}

\textbf{Document: S\~ao Paulo (disambiguation).}
S\~ao Paulo is the capital city of the state of S\~ao Paulo in Brazil. S\~ao Paulo may also refer to other entities.

\vspace{0.45em}

\textbf{Document: S\~ao Jo\~ao Batista.}
S\~ao Jo\~ao Batista or S\~ao Jo\~ao Baptista may refer to several places.

\vspace{0.45em}

\textbf{Document: Campeonato Paulista Estadual de Basquete.}
The Campeonato Paulista Estadual de Basquete is an annual men's professional club basketball competition.

\vspace{0.8em}

\noindent\rule{\linewidth}{0.4pt}

\vspace{0.6em}

\textcolor{black}{\textbf{\large Query:}}
Which film came out earlier, \textbf{13 At A Table} or \textbf{S\~ao Paulo, Sociedade An\^onima}?

\vspace{0.8em}

\colorbox{blue!10}{%
    \parbox{\dimexpr\linewidth-2\fboxsep\relax}{
        \centering
        \textbf{TaC-C}
    }
}

\vspace{0.5em}

\begin{tcolorbox}[
    breakable,
    colback=white,
    colframe=blue!18,
    boxrule=0.5pt,
    arc=1pt,
    left=6pt,
    right=6pt,
    top=5pt,
    bottom=5pt
]
Okay, need to figure out which film came out first between \textit{13 At A Table} and \textit{S\~ao Paulo, Sociedade An\^onima}. Let me check the context provided.
From the documents, \textit{13 at a Table} is a \textbf{2004} film. \textit{S\~ao Paulo, Sociedade An\^onima} is from \textbf{1965}. So the later one is \textit{13 at a Table}, meaning \textit{S\~ao Paulo, Sociedade An\^onima} came out earlier. The key information is the years \textbf{2004} and \textbf{1965}.
\end{tcolorbox}

\colorbox{orange!12}{%
    \parbox{\dimexpr\linewidth-2\fboxsep\relax}{
        \centering
        \textbf{LLMLingua-2}
    }
}

\vspace{0.5em}

\begin{tcolorbox}[
    breakable,
    colback=white,
    colframe=orange!25,
    boxrule=0.5pt,
    arc=1pt,
    left=6pt,
    right=6pt,
    top=5pt,
    bottom=5pt
]
{\ttfamily\small
13 Table Tredici 2004 Italian Enrico Oldoini 2005 Tokyo International Film Festival;
Paulo Sociedade An\^onima 1965 S\'ergio Person Best Foreign Language 38th Academy Awards;
Trolleybuses two SPTrans 1949 Transportes 1988 Naples;
S\~ao;
Governors Paulo;
Chafing dish burner;
S\~ao Paulo under 20 S\~ao Paulo organized Paulista Football Federation traditional final 25 January S\~ao Paulo;
capital;
Campeonato Paulista Estadual Basquete men's.
}
\end{tcolorbox}

\colorbox{red!8}{%
    \parbox{\dimexpr\linewidth-2\fboxsep\relax}{
        \centering
        \textbf{LongLLMLingua}
    }
}

\vspace{0.5em}

\begin{tcolorbox}[
    breakable,
    colback=white,
    colframe=red!22,
    boxrule=0.5pt,
    arc=1pt,
    left=6pt,
    right=6pt,
    top=5pt,
    bottom=5pt
]
{\ttfamily\small
13 at a3 at a Table also known Tredici a tavola'') is a 204 comedy written and directed byrico Oldo. It was entered into competition at the 005 Tokyo International Film.
S\~ao Paulo, Sociade Anima Paulo,adeima a film Luis Person film was S\~ao Paulo State.
}
\end{tcolorbox}

\colorbox{purple!10}{%
    \parbox{\dimexpr\linewidth-2\fboxsep\relax}{
        \centering
        \textbf{PCC}
    }
}

\vspace{0.5em}

\begin{tcolorbox}[
    breakable,
    colback=white,
    colframe=purple!25,
    boxrule=0.5pt,
    arc=1pt,
    left=6pt,
    right=6pt,
    top=5pt,
    bottom=5pt
]
{\ttfamily\small
trolleybuses in two other metropolitan areas, Naples and S\~ao Paulo;
S\~ao Lu\'is de Par\'a or S\~ao Jo\~ao Baptista do Piau\'i the place of the 1954 International Film Festival;
Luis S\'ergio Person's film The Brazilian Selection for Best Foreign Language Feature at the 38th Academy Awards;
Trolleybus network in S\~ao Paulo two systems with one operating under state control and another privately owned;
Copa S\~ao Paulo de Futebol J\'unior a tournament organized by the Paulist\~ao Football Association;
Brazilian football league championship annual men's professional club competition;
What is a trolley system in Greater S\~ao Paulo Greater S\~ao Paulo has its own public transport system comprising greater than half of the city's population;
Where was the first international festival held Tokyo;
Who directed Enoldi Enrico Oldoini;
When did the movie debut 2004;
Which country had more than one subway system Worldwide only one major city---London---is known to have both;
How many teams participated in Copa S\~ao Paulo Futuro-6 until it moved up to an 20--under team squad;
Why are there no women on board chefs' serving dishes;
In which year were they founded 1947;
What does ``casa'' mean home;
Whose portrait appears on Brazil currency Jos.
}
\end{tcolorbox}

\end{tcolorbox}




\end{document}